\documentclass[12pt]{article}
\usepackage[utf8]{inputenc}
\usepackage{hyperref}
\usepackage{amsmath}
\usepackage{graphicx}
\usepackage{enumitem}
\usepackage{tcolorbox}
\usepackage{algorithm}
\usepackage{algorithmic}
\usepackage{subcaption}
\usepackage{amssymb}
\usepackage{amsfonts}
\usepackage{float}

\title{\vspace{-2cm}Generative AI for Industrial Contour Detection: A Language-Guided Vision System\vspace{1cm}}

\author{
    Liang Gong\textsuperscript{1}, Tommy(Zelin) Wang\textsuperscript{1}, Sara Chaker\textsuperscript{1}, Yanchen Dong\textsuperscript{1}, \\
    Fouad Bousetouane\textsuperscript{1}, Brenden Morton\textsuperscript{2}, Mark Mendez\textsuperscript{2} \\[1em]
    \textsuperscript{1}\text{The University of Chicago, USA} \\ 
    \textsuperscript{2}\text{FabTrack, USA} \\[1.5em]
    {\small \texttt{\{gongl, zw1226, sarachaker, yanchendong, bousetouane \}@uchicago.edu}} \\[-0.25em]
    {\small \texttt{\{brenden, mark\}@fabtrack.io}}
}

\date{}

\begin{document}

\maketitle

\begin{abstract}
Industrial computer vision (CV) systems often struggle with noise, material variability, and uncontrolled imaging conditions, limiting the effectiveness of classical edge detectors and handcrafted pipelines. In this work, we present a language-guided generative vision system for remnant contour detection in manufacturing, designed to achieve CAD-level precision. The system is organized into three phases: (1) data preprocessing, (2) contour generation using a conditional GAN, and (3) multimodal contour refinement through vision--language modeling, where standardized prompts are crafted in a human-in-the-loop process and applied through image--text guided synthesis.  

On proprietary FabTrack datasets, the proposed system improved contour fidelity, enhancing edge continuity and geometric alignment while reducing manual tracing. For the refinement phase, we benchmarked several VLMs, including Google’s Gemini 2.0 Flash, OpenAI’s GPT-image-1 integrated within a VLM-guided workflow, and open-source baselines. Under standardized conditions, GPT-image-1 consistently outperformed Gemini 2.0 Flash in both structural accuracy and perceptual quality. These findings demonstrate the promise of VLM-guided generative workflows for advancing industrial CV beyond the limitations of classical pipelines and moving contour detection closer to CAD-level precision.
\end{abstract}

\newpage
\tableofcontents
\newpage

\section{Introduction}

Industrial computer vision (CV) has become a cornerstone of modern manufacturing, supporting applications such as automated inspection, predictive maintenance, and digital twins \cite{fuller2020}. Within these workflows, contour detection plays a critical role in tasks such as remnant reuse, part segmentation, and CAD vectorization. Reliable contour extraction directly improves nesting efficiency, reduces material waste, and lowers the dependence on manual tracing, thereby contributing to both cost savings and sustainability.

Traditional edge-based methods provided the first solutions for contour detection but struggle under the uncontrolled conditions of factory floors. Deep learning brought improvements through models such as Holistically-Nested Edge Detection (HED) \cite{xie2015}, and more recently through large-scale foundation models like the Segment Anything Model (SAM) \cite{kirillov2023} and DINO \cite{caron2021}. While these approaches generalize well across diverse domains, they often produce incomplete or unstable contours when applied to reflective surfaces, irregular geometries, and variable lighting common in industrial settings.

Generative AI (GenAI) introduces a new paradigm by moving beyond detection toward transformation and refinement. Generative Adversarial Networks (GANs) \cite{goodfellow2014} established the foundation for high-fidelity synthesis, followed by image-to-image translation methods such as Pix2Pix \cite{isola2017} and CycleGAN \cite{zhu2017}, which demonstrated practical domain adaptation. More recently, diffusion models \cite{ho2020,dhariwal2021,rombach2022} have further advanced stability and visual quality. In parallel, multimodal vision--language models (VLMs) \cite{radford2021,chen2025} enable contour refinement guided by natural-language instructions, opening the door to interactive and adaptive manufacturing workflows. Together, these advances suggest the possibility of systems that transform noisy, uncontrolled inputs into CAD-precision representations suitable for engineering use.

\subsection{Motivation and Scope}

The manufacturing sector is being reshaped by the Industry 4.0 paradigm, where automation, digitization, and sustainability are strategic priorities. Companies must balance flexible production and material efficiency while addressing the shortage of skilled labor for CAD digitization. Existing CV solutions, while valuable for inspection and simulation \cite{fuller2020}, typically assume controlled imaging conditions, making them unreliable on the factory floor where materials, lighting, and setups vary widely.

Generative AI provides a promising alternative by enabling the translation, refinement, and regularization of noisy imagery. In this work, we adapt these advances to the challenge of \emph{remnant digitization}—the process of extracting usable geometric information from leftover raw material sheets (remnants) after initial cutting operations. Accurate digitization of remnants enables manufacturers to reuse off-cuts in subsequent jobs, contributing directly to efficiency, waste reduction, and operational scalability. By focusing on this problem, we aim to demonstrate how language-guided generative systems can extend industrial CV capabilities under real-world, uncontrolled conditions.

\subsection{Problem Statement}

Despite progress in computer vision, remnant contour extraction remains a bottleneck in manufacturing workflows. Leftover sheets of metal, wood, or plastic must be captured and converted into CAD-ready contours before reuse. Current workflows face several challenges:
\begin{itemize}
    \item Manual tracing is slow, inconsistent, and costly at scale.
    \item Complex geometries with low-contrast edges, reflective surfaces, and irregular shapes complicate accurate processing.
    \item Existing contour models, including advanced foundation models such as HED \cite{xie2015}, SAM \cite{kirillov2023}, and DINO \cite{caron2021}, often fail to produce stable, production-ready contours in these settings.
    \item Commercial nesting software assumes clean CAD inputs, rendering raw images unsuitable without extensive preprocessing.
\end{itemize}

\begin{figure}[H]
    \centering
    \includegraphics[width=0.9\linewidth]{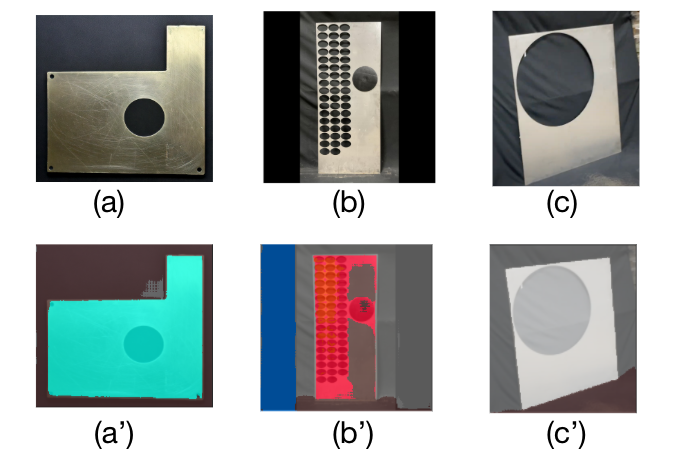}
    \caption{Examples of remnant images captured at the FabTrack Lab (a,b,c) and corresponding zero-shot mask predictions generated using DINOv2 (a',b',c').}
    \label{fig:remnant_dino}
\end{figure}

Figure~\ref{fig:remnant_dino} illustrates this challenge. The top row shows raw remnant photographs captured at the FabTrack Lab, while the bottom row shows zero-shot masks generated by DINOv2 \cite{oquab2023dinov2}. Although the model can roughly localize material regions, the masks suffer from blurred edges, missing details, and irregular contours. These shortcomings highlight the limitations of general-purpose segmentation models in industrial contexts, where CAD-level precision is essential for downstream reuse and digitization.

\subsection{Contributions}

The main contributions of this work are as follows:
\begin{itemize}
    \item We propose a three-phase language-guided generative vision system for industrial remnant contour detection, achieving CAD-level precision under real-world factory conditions.  
    \item We adapt and fine-tune a conditional GAN (Pix2Pix) for industrial remnant contour generation, producing denoised and continuous edge maps that serve as reliable inputs for downstream contour refinement and CAD-oriented workflows.  
    \item We integrate multimodal contour refinement within a VLM-guided workflow, where standardized natural-language prompts and human-in-the-loop adjustments enable geometric regularization and improved alignment.  
    \item We introduce an industrial evaluation protocol that combines quantitative image similarity metrics with operator feedback, ensuring both perceptual quality and practical usability.  

\end{itemize}

\section{Background and Prior Work}

\begin{figure}[H]
    \centering
    \includegraphics[width=0.9\linewidth]{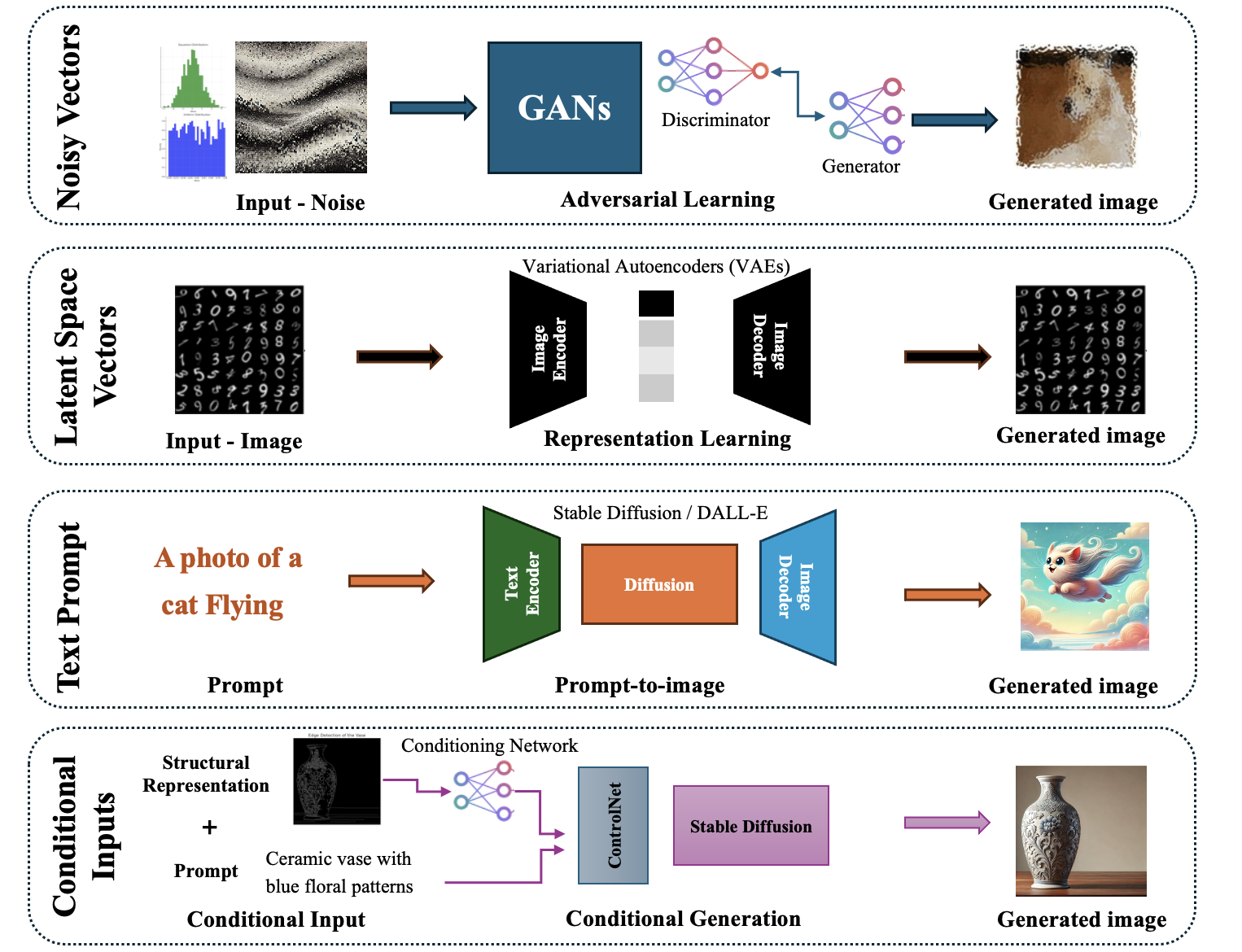}
    \caption{Four-pillar taxonomy of input-driven generative AI for vision, reprinted from \cite{bousetouane2025}.}
    \label{fig:taxonomy}
\end{figure}

Recent advances in generative AI for vision can be systematically categorized using the four-pillar taxonomy introduced by Bousetouane \cite{bousetouane2025}. This taxonomy, illustrated in Figure~\ref{fig:taxonomy}, frames generative models around the nature of their inputs: (i) noisy vectors, (ii) latent-space vectors, (iii) text prompts, and (iv) conditional inputs. This input-centric perspective provides a foundation for understanding the diverse architectures and workflows that enable image synthesis, translation, and refinement. Each pillar represents a distinct family of generative techniques with specific trade-offs in fidelity, controllability, and scalability.

\subsection{Noisy Vectors: GANs and Diffusion Models}

Generative Adversarial Networks (GANs) \cite{goodfellow2014} established adversarial training between a generator and discriminator, producing high-fidelity images from random noise. Conditional frameworks such as pix2pix \cite{isola2017} \cite{isola2017} and CycleGAN \cite{zhu2017} extended this paradigm to paired and unpaired image-to-image translation. StyleGAN \cite{karras2019} introduced a style-based latent control mechanism that achieved unprecedented realism and fine-grained manipulation.

In parallel, diffusion models emerged as a powerful alternative. The Denoising Diffusion Probabilistic Model (DDPM) \cite{ho2020} proposed iterative denoising of Gaussian noise, followed by guided diffusion \cite{dhariwal2021} that enabled controllable synthesis. Latent diffusion, implemented in Stable Diffusion \cite{rombach2022}, significantly reduced computational costs while delivering high-resolution, diverse outputs. These models now underpin state-of-the-art visual generation pipelines and are increasingly adapted for industrial imaging tasks.

\subsection{Latent-Space Vectors: Variational Autoencoders}

Variational Autoencoders (VAEs), introduced by Kingma and Welling \cite{kingma2013}, encode data into a probabilistic latent space and reconstruct it back. This framework balances reconstruction fidelity with latent regularization. Subsequent extensions include $\beta$-VAE \cite{higgins2017}, which improved disentanglement of latent factors, Conditional VAE \cite{sohn2015}, which introduced auxiliary variables for guided synthesis, and VQ-VAE \cite{oord2017}, which leveraged discrete codebooks for scalable high-fidelity synthesis. VAEs are particularly valuable for applications that require structured latent spaces, such as anomaly detection and representation learning, but are less dominant in high-quality image synthesis compared to GANs and diffusion.

\subsection{Text Prompts: Prompt-to-Image Pipelines}

The advent of prompt-to-image generation transformed generative AI into an accessible and controllable tool. Early text encoders such as BERT \cite{devlin2019} and GPT \cite{radford2019} demonstrated the ability to capture semantic information from natural language. Multimodal encoders like CLIP \cite{radford2021} and its extensions (e.g., FLAVA, CoCa) aligned text and vision into joint embeddings, enabling powerful text-to-image pipelines. Stable Diffusion \cite{rombach2022} combines diffusion modeling with latent-space learning for efficient high-resolution generation. Autoregressive frameworks such as DALL·E \cite{ramesh2022} and more recent unified multimodal systems like DeepSeek Janus-Pro \cite{chen2025} extend this capability to complex compositional reasoning. These systems illustrate the growing role of text prompts as flexible, human-friendly inputs for controlling generative models.

\subsection{Conditional Inputs: Structure-Guided Generation}

Beyond noise, latents, and text, conditional inputs provide structural control over generative outputs. ControlNet \cite{zhang2023controlnet} demonstrated how side networks can guide diffusion models using edge maps, depth, or human poses. T2I-Adapter \cite{gao2023t2iadapter} introduced lightweight adapters to condition on layout, style, or depth. Pose-Guided GANs \cite{ma2017posegan} allowed targeted synthesis of human poses, while compositional diffusion frameworks \cite{liu2023compositional} combined multiple conditions to achieve semantically consistent outputs. These methods highlight how conditional generative modeling can produce CAD-ready or structurally aligned results, which is particularly relevant for industrial applications such as contour generation.

This body of work illustrates a rapid progression from early noise-driven synthesis toward controllable, multimodal, and structure-aware generative systems \cite{bousetouane2025}. Each pillar of the taxonomy offers unique advantages and limitations when considered for industrial contour detection. GANs \cite{goodfellow2014} and diffusion models \cite{ho2020,dhariwal2021,rombach2022} are capable of producing high-fidelity outputs, yet they often lack the geometric precision needed for engineering-grade contours. VAEs \cite{kingma2013,higgins2017,sohn2015} provide interpretable latent spaces that are valuable for representation learning, but their reconstructions typically fall short in visual sharpness. Prompt-to-image pipelines \cite{radford2021,ramesh2022,chen2025} bring flexibility and natural language control, although they remain better suited for creative tasks than for CAD-level accuracy. Conditional approaches such as ControlNet \cite{zhang2023controlnet}, T2I-Adapter \cite{gao2023t2iadapter}, and compositional diffusion \cite{liu2023compositional} represent the most promising direction, since they explicitly integrate structural guidance into the generation process, aligning outputs with downstream CAD and manufacturing requirements. Taken together, these advances motivate the development of hybrid generative pipelines that combine fidelity, controllability, and structure-awareness to meet the stringent demands of industrial digitization workflows.

\section{Proposed System}
\label{sec:proposed}

\subsection{Overview}

\begin{figure}[H]
    \centering
    \includegraphics[width=1\linewidth]{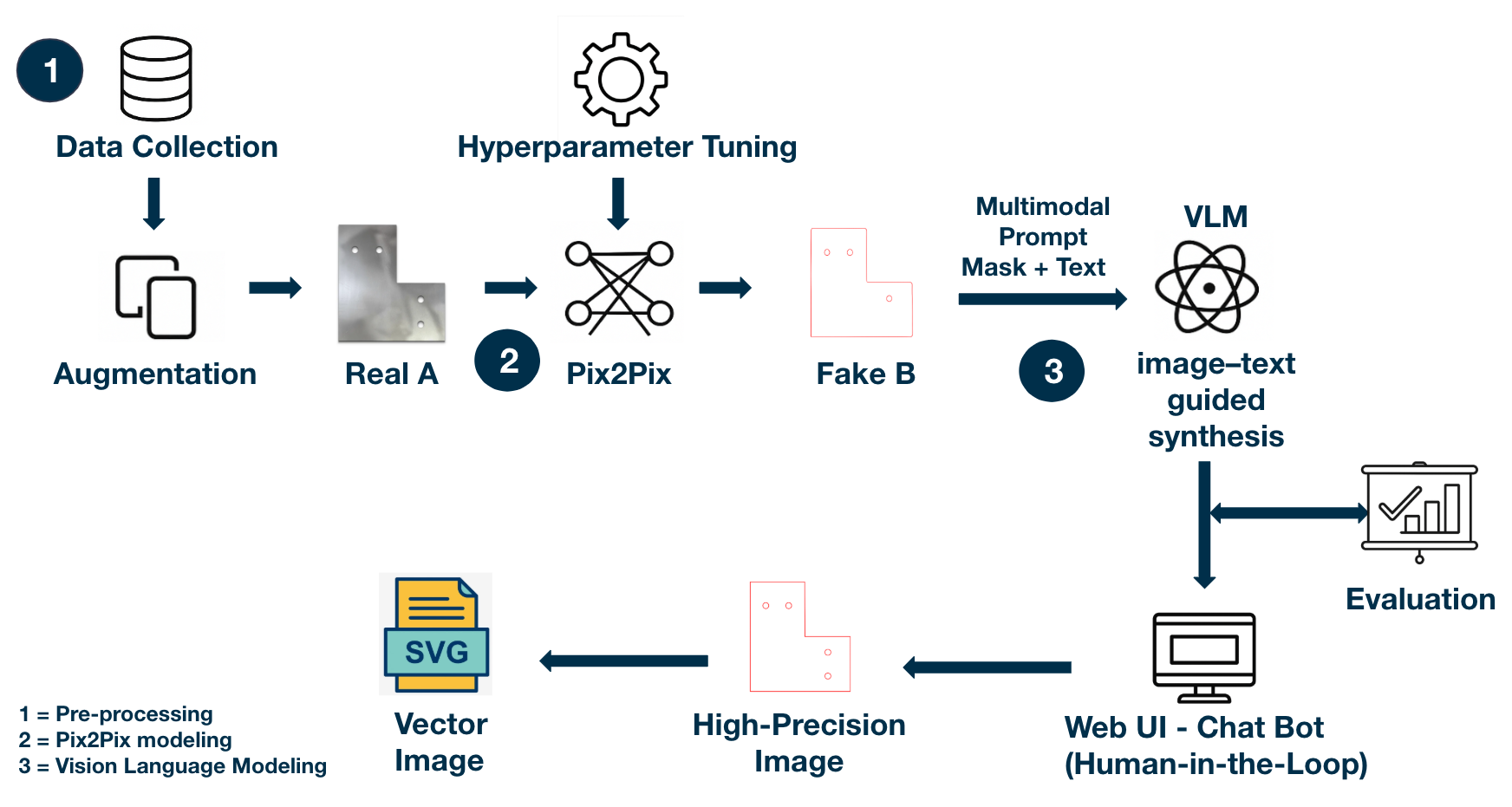}
    \caption{Overview of the proposed three-phase generative system for industrial contour refinement, combining preprocessing, Pix2Pix contour generation, and multimodal prompt-guided refinement with human-in-the-loop support.}
    \label{fig:system_pipeline}
\end{figure}

We propose an end-to-end generative system for robust remnant contour extraction in industrial manufacturing. The system integrates three main phases: (1) preprocessing and data augmentation, (2) contour generation using Pix2Pix \cite{isola2017}, and (3) contour refinement through vision--language modeling, where the Pix2Pix output image is combined with standardized text instructions and applied through an image generation model that supports multimodal prompting (image + text). In this setup, GPT-image-1 \cite{gptimage2024} serves as the refinement backend within a VLM-guided workflow, demonstrating how multimodal prompting can drive precise contour correction. A web-based chatbot interface enables human-in-the-loop refinement by allowing experts to iteratively adjust prompts and review outputs. The complete system is illustrated in Figure~\ref{fig:system_pipeline}.

\subsection{Data Acquisition and Preprocessing (Phase 1)}

To capture the variability present in real manufacturing environments, remnant images were collected from FabTrack’s industrial sites, representing diverse materials, shapes, scales, and lighting conditions. All images were standardized to $1024 \times 1024$ resolution via center-cropping or zero-padding and normalized to RGB format to ensure consistency across the dataset. 

Dynamic augmentations were applied during training, including horizontal flips, random rotations, and brightness shifts, to simulate natural variability and improve generalization. This preprocessing pipeline ensured stable training behavior and compatibility with downstream architectures.

\subsection{Contour Generation with pix2pix  (Phase 2)}

The first modeling stage employed pix2pix \cite{isola2017}, a conditional GAN architecture for paired image-to-image translation. The model was trained on a hybrid dataset composed of synthetic pairs generated programmatically and real-world samples curated and annotated by FabTrack domain experts. This mixture allowed the model to learn both structural regularities from synthetic data and the complexities of industrial imagery from real examples. 

Hyperparameter tuning was carried out extensively, including adjustments to the learning rate, L1 loss weighting, GAN mode, generator depth, normalization strategies, and batch size. The U-Net generator and PatchGAN discriminator were adopted as core components, consistent with prior work in image-to-image translation. The result of this stage was the generation of binary contour masks that were coherent and denoised but not yet sufficiently precise for CAD integration.

\subsection{VLM-based Contour Refinement (Phase 3)}

The third phase of the system focused on contour refinement through vision--language modeling. Here, the Pix2Pix \cite{isola2017} output masks were paired with standardized text instructions to form multimodal prompts. These prompts were passed to GPT-image-1 \cite{gptimage2024}, an image generation model that supports multimodal prompting (image + text), which produced refined contours with improved geometric fidelity. This phase provided the critical step toward achieving CAD-level precision in the contours, addressing the structural limitations of Pix2Pix alone.

\subsection{Human-in-the-Loop Chatbot Interface}

To make refinement more accessible to domain experts, a web-based chatbot interface was integrated into the workflow. Instead of crafting detailed technical prompts, operators could issue natural-language requests (e.g., “remove noise in the top-right corner,” “make all holes uniform”). These requests were translated into structured prompts and applied in real time, allowing experts to iteratively guide the refinement process. The interface did not add a new modeling component but served as a practical mechanism for embedding expert oversight into the VLM-guided workflow, ensuring that outputs remained operationally relevant and achieved CAD-level precision.  

\subsection{Alternative Models Considered}

Several alternative multimodal models were evaluated during development. Gemini \cite{google2025gemini}, though strong at multimodal reasoning, often introduced geometric distortions and hallucinated cutouts when applied to structural contour refinement. DeepSeek-VL \cite{deepseek2025} and Qwen-VL \cite{qwen2023} demonstrated competence in image--text alignment but lacked the editing functionality required for this task. Moondream \cite{moondream2024}, while efficient and lightweight, struggled to maintain fine structural fidelity. These limitations highlighted the strengths of GPT-image-1, which consistently delivered more reliable contour refinements within the standardized VLM-guided workflow. Detailed quantitative comparisons are presented in Section~\ref{sec:Evaluation}.

\newpage

\section{Results and Analysis}

\subsection{Dataset and Experimental Setup}

All experiments were conducted on proprietary data provided by FabTrack, consisting of remnant images captured directly from real manufacturing sites. This dataset reflects authentic industrial variability in materials, scales, and imaging conditions. To improve generalization and enable fine-tuning of pix2pix \cite{isola2017}, we applied dynamic augmentations (flips, rotations, brightness shifts) and upsampled the dataset. Hybrid training pairs were created by combining synthetic samples with real-world images that were manually curated and annotated by FabTrack experts.

\subsection{Finetuned pix2pix  model  Evaluation}

The fine-tuned pix2pix \cite{isola2017} model achieved strong perceptual similarity across evaluation metrics. Structural Similarity Index (SSIM) \cite{wang2004ssim} reached \textbf{0.9705}, and LPIPS \cite{zhang2018lpips} scored a low \textbf{0.0530}, indicating that generated contours were visually close to ground truth. However, geometric fidelity showed moderate errors, with a Hausdorff Mean Distance \cite{huttenlocher1993comparing} of \textbf{58.22} and an IoU of \textbf{0.6265}. These results confirm that pix2pix \cite{isola2017} excels in generating visually coherent contours but struggles with fine-grained geometric accuracy.

\begin{table}[H]
\centering
\caption{Evaluation of the fine-tuned Pix2Pix model on the FabTrack test set}
\label{tab:pix2pix}
\begin{tabular}{lcc}
\hline
\textbf{Metric} & \textbf{Value} & \textbf{Interpretation} \\
\hline
SSIM             & 0.9705 & High structural similarity \\
LPIPS            & 0.0530 & Low perceptual distortion \\
Hausdorff Distance & 58.22  & Moderate geometric drift \\
IoU              & 0.6265 & Good region overlap \\
\hline
\end{tabular}
\end{table}

\subsection{VLM Evaluation on Real FabTrack Images}
\label{sec:Evaluation}

For evaluation, we compared contour refinement across APIs under identical conditions. Both OpenAI’s \texttt{gpt-image-1} endpoint \cite{gptimage2024} and Google’s \texttt{gemini-2.0-flash-preview-image-generation} API \cite{google2025gemini} were called using the same standardized prompt and the Pix2Pix \cite{isola2017} contour masks as inputs.  

All results were generated via API calls with identical inputs and consistent evaluation metrics, ensuring a fair and reproducible comparison. Quantitative results are reported in Table~\ref{tab:vlm}.

\begin{table}[H]
\centering
\caption{Comparison of VLM-Guided Refinement on FabTrack Real-Image Test Subset}
\label{tab:vlm}
\resizebox{\textwidth}{!}{%
\begin{tabular}{lcc}
\hline
\textbf{Metric} & \textbf{OpenAI API (GPT-image-1, n=274)} & \textbf{Google API (Gemini 2.0 Flash, n=274)} \\
\hline
SSIM & 0.9044 & 0.6525 \\
LPIPS & 0.2315 & 0.3154 \\
Hausdorff Dist. & 98.046 & 161.353 \\
IoU & 0.6420 & 0.6505 \\
\hline
\end{tabular}%
}
\end{table}

The results in Table~\ref{tab:vlm} show that OpenAI’s GPT-image-1 consistently outperformed Google’s Gemini 2.0 Flash API under identical testing conditions. GPT-image-1 achieved higher SSIM and lower LPIPS values, indicating closer perceptual alignment with ground truth and reduced visual distortion. It also produced a substantially lower Hausdorff distance, reflecting improved geometric fidelity in contour reconstruction. While IoU scores were broadly comparable, GPT-image-1 was more effective at preserving fine structural details, whereas Gemini 2.0 Flash exhibited greater variability in alignment. Overall, these results suggest that the Flash API, despite Gemini’s broader multimodal reasoning strengths, is less well suited for specialized contour refinement tasks compared to OpenAI’s pipeline.

\subsection{Qualitative Comparison of Contour Refinement}

\begin{figure}[H]
    \centering
    \includegraphics[width=1\linewidth]{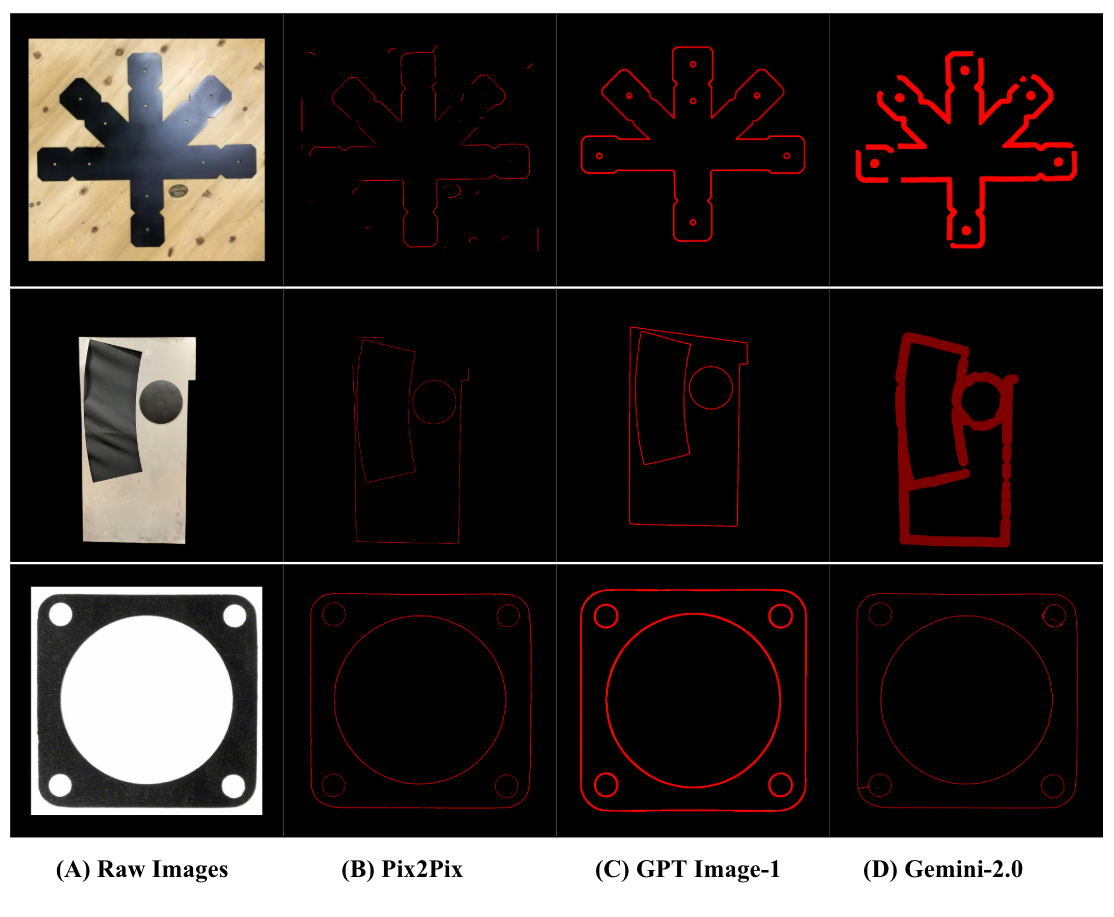}
    \caption{Qualitative comparison of contour refinement methods.}
    \label{fig:qualitative_examples}
\end{figure}

\begin{figure}[H]
    \centering
    \includegraphics[width=0.63\linewidth]{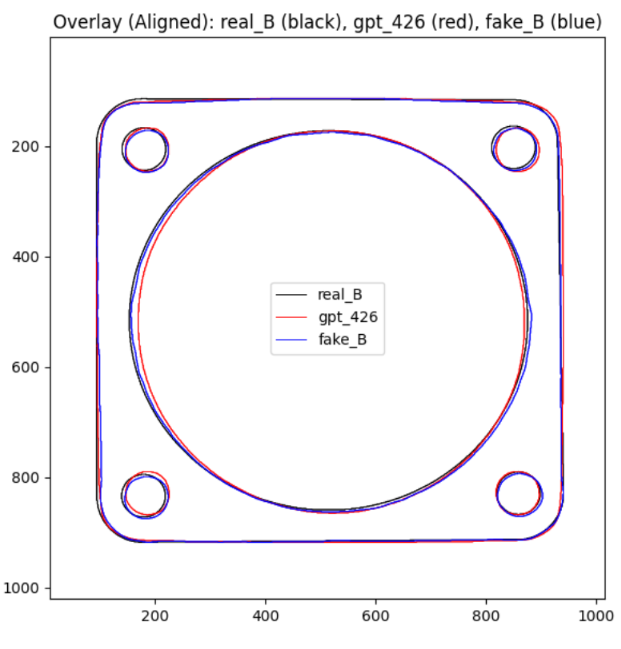}
    
    \vspace{0.5em} 
    
    \includegraphics[width=0.7\linewidth]{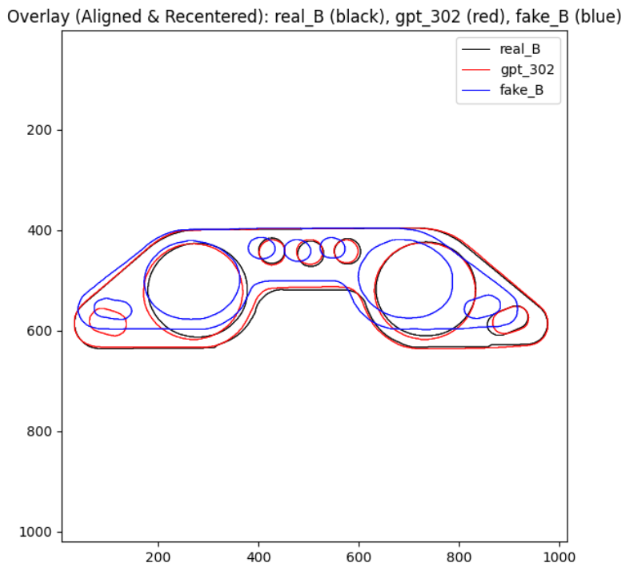}
    
    \caption{Overlay visualization of contour alignment. 
    Top: Ground truth (Real B) shown in black, pix2pix \cite{isola2017} output (Fake B) in blue, and GPT-image-1 enhanced contours in red. 
    Bottom: Additional comparison highlighting how prompt-guided VLM refinement not only filled missing contour gaps but also realigned contours to closely match the ground truth, resulting in higher geometric accuracy and reliability for CAD integration.}
    \label{fig:overlay_examples}
\end{figure}

Quantitative results were complemented with qualitative comparisons to better illustrate the role of VLM refinement in improving contour fidelity. Representative examples from the FabTrack dataset are shown in Figures~\ref{fig:qualitative_examples} and \ref{fig:overlay_examples}.

Column (A) shows raw input remnant images. 
Column (B) presents contours generated by the fine-tuned Pix2Pix model \cite{isola2017}. 
Column (C) displays refined contours produced with OpenAI’s GPT-image-1 using the standardized multimodal prompt. 
Column (D) illustrates refined contours generated with Google’s Gemini 2.0 Flash API. 
GPT-image-1 outputs exhibit sharper, more consistent boundaries aligned with object geometry, while Gemini 2.0 Flash often introduces variability or distortions.

\section{Conclusion and Future Directions}

This work introduced a three-phase generative system for industrial remnant contour extraction. Phase~1 focused on data preprocessing and augmentation, Phase~2 applied a Pix2Pix-based conditional GAN for contour generation, and Phase~3 integrated vision--language modeling for refinement using multimodal prompts. Leveraging GPT-image-1 within a prompt-guided refinement loop and a human-in-the-loop chatbot interface, we addressed the geometric limitations of Pix2Pix, producing contours that were sharper, more consistent, and better aligned with ground-truth references. While several VLMs were explored during development, GPT-image-1 was selected as the most effective component of the proposed system.  

The experimental analysis confirms that the proposed three-phase generative system can significantly reduce manual tracing effort, improve accuracy, and generate contours with CAD-level precision suitable for downstream manufacturing workflows. Visual overlays and metric evaluations demonstrated that VLM-enhanced refinement not only corrected structural drift but also improved local alignment in complex geometries, bridging the gap between automated contour generation and industrial-grade requirements.  

Looking ahead, there remain several promising directions for future research. First, expanding the dataset with additional industrial modalities (e.g., reflective alloys, composite materials) would further validate robustness across domains. Second, incorporating active learning loops could allow the system to continuously adapt to new environments with minimal expert intervention. Finally, a key avenue lies in extending this work toward \textit{agentic multimodal AI systems}, as articulated in recent research on Agentic Systems \cite{bousetouane2025agentic} and Physical AI Agents \cite{bousetouane2025physical}. Such systems, capable of autonomously orchestrating preprocessing, contour generation, and refinement, hold the potential to manage more complex vision tasks and dynamically adapt to diverse imaging conditions. These advances will pave the way for higher autonomy in manufacturing vision systems, accelerating progress toward Industry 4.0 digitization goals.

\newpage

\section{Acknowledgment}
We thank FabTrack for data and domain guidance, the University of Chicago MSADS program for research support, and Dr.Fouad Bousetouane for his coaching and guidance.

\bibliographystyle{plain}
\bibliography{science_template}

\end{document}